\documentclass[runningheads,a4paper]{llncs}
\usepackage{amsmath}
\usepackage{amsfonts}
\usepackage{amssymb}
\usepackage{graphicx}
\usepackage{enumerate}
\usepackage[capitalise]{cleveref}
\crefname{section}{Sect.}{Sect.}
\crefname{equation}{}{}
\crefname{algorithm}{Alg.}{Alg.}
\usepackage{bm}
\usepackage{url}
\usepackage{tikz}
\usetikzlibrary{positioning}
\usetikzlibrary{shapes.geometric}
\usepackage{circuitikz}
\usepackage{algorithm}
\usepackage{algpseudocode}
\algnewcommand\algorithmicinput{\textbf{Input:}}
\algnewcommand\Input{\item[\algorithmicinput]}
\algnewcommand\algorithmicoutput{\textbf{Output:}}
\algnewcommand\Output{\item[\algorithmicoutput]}
\algnewcommand\algorithmicforeach{\textbf{for each}}
\algdef{S}[FOR]{ForEach}[1]{\algorithmicforeach\ #1\ \algorithmicdo}

\makeatletter
\AtBeginDocument{%
  \def\doi#1{\url{https://doi.org/#1}}}
\makeatother

\newcommand{\R}[0]{\mathbb{R}}
\newcommand{\C}[0]{\mathbb{C}}
\newcommand{\Q}[0]{\mathbb{Q}}

\newcommand{\cF}[0]{\mathcal{F}}

\newcommand{\cS}[0]{\mathcal{S}}

\newcommand{\p}{\bm{p}}
\newcommand{\rr}{\bm{r}}

\newcommand{\cR}{\mathcal{R}}

\begin{document}
\title{
  Trajectory Planning and Certification for 3-DOF Robot Manipulators Using Real Quantifier Elimination Based on Comprehensive Gr\"obner Systems
}
\titlerunning{Trajectory Planning and Certification for 3-DOF Robot Manipulators}
\author{
    Yu Nakai\inst{1}\and
    Akira Terui\inst{1}\orcidID{0000-0003-0846-3643} \and
    Masahiko Mikawa\inst{1}\orcidID{0000-0002-2193-3198}
}
\institute{%
University of Tsukuba, Tsukuba, Japan \\
\email{terui@math.tsukuba.ac.jp}\\
\email{mikawa@slis.tsukuba.ac.jp}\\
\url{https://researchmap.jp/aterui}
}

\maketitle

\begin{abstract}
    We propose an algorithm and its implementation for trajectory planning and certification for 3-DOF robot manipulators. The method uses Real Quantifier Elimination (QE) based on Comprehensive Gröbner Systems (CGS), also known as the CGS-QE method.  
    The main advantage of the proposed method is its efficiency in trajectory planning and solution certification. This efficiency comes from the effective use of the CGS. 
    First, for trajectory planning, we solve the inverse kinematics problem at each point along the trajectory via Gröbner basis computation. This usually requires recalculating the Gröbner basis at every point, which is time-consuming. We avoid this by computing the CGS for a parametric system. Here, the end-effector coordinates are parameters. This approach streamlines the algorithm.
    Second, for solution certification, the CGS-QE method certifies that an inverse kinematics solution exists at any point along the end-effector's trajectory. Our method also certifies solutions for trajectories composed of line segments and cubic natural splines.
    The algorithm is implemented within the computer algebra system Risa/Asir, and experimental results are presented.
    \keywords{Comprehensive Gr\"obner Systems \and Robotics \and Inverse kinematics \and
    Trajectory planning}
\end{abstract}

\section{Introduction}
\label{sec:introduction}

In this paper, we discuss the motion planning of a 
robot manipulator. 
A manipulator is a robot that consists of a series of links connected by joints like a human arm.
The last link connected to the end-effector.
We deal with a manipulator 
``myCobot 280'' \cite{mycobot-280}
(hereafter, we refer to it as ``myCobot''),
which has six revolute joints and therefore is a 6-DOF manipulator.

Motion planning involves solving the inverse kinematics and trajectory planning problems.
Inverse kinematics finds a joint configuration for a given end-effector position and orientation.
Trajectory planning extends this problem to a continuous end-effector path.

In computer algebra, the inverse kinematics problem is solved by reducing it to a system of polynomial equations and then computing a Gröbner basis 
(see \cite{fau-mer-rou2006,kal-kal1993,ric-sch-ces2021} and the references therein).
The trajectory planning problem is solved by applying inverse kinematics at each point along the trajectory.

An advantage of using Gröbner bases is that we can find global solutions, which allows the  feasibility of the end-effector's motion to be determined before actual execution, unlike numerical methods.
However, Gröbner basis computation can be more computationally intensive than numerical methods, and repeated computation along the trajectory can be time-consuming.

Our research group has proposed methods for solving inverse kinematics and trajectory planning problems for manipulators using Gröbner bases 
\cite{oka-ter-mik2025b,oka-ter-mik2025a,ota-ter-mik2021,shi-oka-ter-mik2024,yos-ter-mik2023}.
A core feature of our proposed method is the use of 
Comprehensive Gröbner Systems (CGS) \cite{suz-sat2006,wei1992} to avoid the repeated computation of Gröbner bases by precomputing the CGS for a parametric system of equations, thereby significantly reducing computational cost.

In this paper, we focus on the CGS-based Quantifier Elimination (CGS-QE) method 
\cite{fuk-iwa-sat2015}, another application of CGS, for certifying solutions to trajectory planning problems.
Our research group's previous studies applied CGS-QE to inverse kinematics and trajectory planning for 3-DOF manipulators \cite{ota-ter-mik2021,yos-ter-mik2023}, guaranteeing solution existence along a trajectory.
A core feature of our method is that, by representing points on a given trajectory, it guarantees the existence of solutions to the inverse kinematic problem rigorously for any point on the trajectory, unlike numerical methods.

Our contribution in this paper is as follows:
\begin{enumerate}
    \item We propose a method for certifying solutions to the trajectory planning problem for myCobot using the CGS-QE method. While our previous work \cite{oka-ter-mik2025b,oka-ter-mik2025a} addressed the trajectory planning problem for myCobot as a 6-DOF manipulator with the end-effector's orientation fixed, here we treat myCobot as a 3-DOF manipulator,  
    considering only the position of the end-effector for computational efficiency, and propose a corresponding method.
    \item In our previous method \cite{yos-ter-mik2023}, the trajectory was given as a line segment; however, in this paper, we propose a solution for a trajectory given as a cubic natural spline, which is a trajectory that enhances the end-effector's ability to move around obstacles.
    \item While our previous method only determined whether trajectory planning was feasible for a given trajectory as a whole, in this paper, we specifically calculate the range of parameters within the given trajectory for which the trajectory planning is feasible.
    \item Previously, we used Wolfram Mathematica for some QE steps. 
    In this paper, we have implemented the entire method with our original algorithm (\cref{alg:detecting-range}) using the computer algebra system Risa/Asir, which is free software.
\end{enumerate}

This paper is organized as follows.
In \cref{sec:inverse-kinematics-problem}, we describe the inverse kinematics computation of myCobot as a 3-DOF manipulator.
In \cref{sec:trajectory-planning-problem}, we propose a method for the trajectory planning problem for myCobot.
In \cref{sec:certification-problem}, we propose a method for certifying solutions to the trajectory planning problem for myCobot using the CGS-QE method.

\section{The Inverse Kinematics Problem of myCobot as a 3-DOF Manipulator}
\label{sec:inverse-kinematics-problem}

In this section, we describe the inverse kinematics (IK) problem of myCobot as a 3-DOF manipulator, following the formulation by Okazaki et al. \cite{oka-ter-mik2025a}\footnote{Note that, although we have already proposed a formulation in Shirato et al. \cite{shi-oka-ter-mik2024}, the formulation in the present paper is different.}.
We formulate the IK problem using the Denavit-Hartenberg (D-H) convention \cite{den-har1955}.

Number each joint of myCobot from 1 to 7, starting from the ground up to the end-effector.  
(note that Joint 7 is the end-effector). 
Let $\Sigma_i$ be the coordinate system w.r.t.\ Joint $i$, and ${}^ix,{}^iy$,
and ${}^iz$ be
the $x$, $y$, and $z$ axes of $\Sigma_i$, respectively.
Let $\mathcal{O}_i$ be the origin of $\Sigma_i$, and
${}^{i-1}l_i$ be the common normal of the axes ${}^{i-1}z$ and ${}^iz$.
According to the D-H convention, for $i=1,\dots,7$, 
$\Sigma_i$ is defined
as a right-handed coordinate system that satisfies the following:
the origin $\mathcal{O}_i$ is placed at Joint $i$;
the axis ${}^iz$ is aligned with the rotation axis of Joint $i$, with the positive
direction pointing towards Joint $i+1$;
the ${}^ix$ and ${}^iy$ axes follow the conventions of the 3D visualization tool
RViz \cite{Rviz}.
Note that $\Sigma_1$ is treated as the global coordinate system.

The transformation matrix ${}^iT_{i+1}$ \cite[(1)]{oka-ter-mik2025a} from $\Sigma_{i-1}$ to $\Sigma_i$ is defined with the following paprameters: the length $a_i$ of the common normal ${}^{i}l_{i+1}$, 
the rotation angle $\alpha_i$ between the ${}^iz$ and
the ${}^{i+1}z$ axes around the ${}^{i+1}x$ axis,
the distance $d_i$ between the common normal ${}^{i}l_{i+1}$ and $\mathcal{O}_i$,
and the rotation angle $\theta_i$ between the common normal
${}^{i}l_{i+1}$ and the ${}^ix$ axis around the ${}^iz$ axis.
Furthermore, the transformation matrix $A_i$ \cite[(2)]{oka-ter-mik2025a} from $\Sigma_{i-1}$ to $\Sigma_i$ is defined 
by adding the rotation angle $\delta_i$ between the ${}^ix$ and the ${}^{i+1}x$ axes 
around the ${}^{i+1}z$ axis according to the coordinate transformation of myCobot in RViz.
Let $A$ be the transformation matrix
from the end-effector's coordinate system $\Sigma_7$ to the global coordinate system $\Sigma_1$.
Then, $A$ is expressed as
  $A=A_1A_2A_3A_4A_5A_6$.

Let $\p={}^t(p_1,p_2,p_3)$ be the position of the end-effector w.r.t.\ $\Sigma_1$.
Since ${}^t(0,0,0)$ is the position of the end-effector w.r.t.\ $\Sigma_7$, we have
\begin{equation}
  \label{eq:end-effector-position}
  {}^t(p_1,p_2,p_3,1)=A\;{}^t(0,0,0,1).
\end{equation}
Furthermore, since the position of the end-effector is determined by Joints 1, 2, and 3, and we ignore the orientation of the end-effector, we assume that $\theta_4=\theta_5=\theta_6=0$.
By putting these values into $A$ and equating the first three rows of 
\eqref{eq:end-effector-position}, we obtain a system of equations w.r.t.\ $\theta_1$, $\theta_2$, and $\theta_3$.
The system of equations is reduced to a system of polynomial equations by substituting 
$c_i=\cos\theta_i$ and $s_i=\sin\theta_i$ for $i=1,2,3$ and using the trigonometric identities $c_i^2+s_i^2=1$ for $i=1,2,3$.
As a result, we formulate the IK problem of myCobot as a 3-DOF manipulator as a system of polynomial equations $f_1=\cdots=f_7=0$ w.r.t.\ $c_1$, $s_1$, $c_2$, $s_2$, $c_3$, and $s_3$ with parameters $p_1$, $p_2$, and $p_3$, where
\begin{equation}
  \label{eq:inverse-kinematics-polynomial-system}
  \begin{split}
    f_1 &=64.62s_{1} + (-48.6c_{1}s_{3} + (-169.18c_{3} - 110.4)c_{1})s_{2} - 169.18c_{2}c_{1}s_{3} \\
    &\quad + 48.6c_{3}c_{2}c_{1} - p_{1}, \\
    f_2 &=((-48.6s_{3} - 169.18c_{3} - 110.4)s_{2} - 169.18c_{2}s_{3} + 48.6c_{3}c_{2})s_{1} \\
    &\quad - 64.62c_{1} - p_{2}, \\
    f_3 &=(-169.18s_{3} + 48.6c_{3})s_{2} + 48.6c_{2}s_{3} + (169.18c_{3} + 110.4)c_{2} \\
    &\quad + 131.56 - p_{3}, \\
    f_4 &=s_{1}^2 + c_{1}^2 - 1, \quad
    f_5 = s_{2}^2 + c_{2}^2 - 1, \quad
    f_6 = s_{3}^2 + c_{3}^2 - 1.
  \end{split}
\end{equation}

To solve \eqref{eq:inverse-kinematics-polynomial-system}, we compute the Gröbner basis for the ideal 
$\langle f_1,\dots,f_6\rangle $ w.r.t.\ the lexicographic order on the variables, for example, 
$c_1>s_1>c_2>s_2>c_3>s_3$ after substituting the parameters $p_1$, $p_2$, and $p_3$.
However, computing the Gröbner basis each time after substituting the parameters is time-consuming.
By computing the CGS for the ideal $\langle f_1,\dots,f_6\rangle $ with parameters $p_1,p_2,p_3$ before operating the manipulator, the corresponding Grobner basis can be obtained by substituting the parameters into the CGS, which is expected to significantly reduce the computing time.

\section{Trajectory Planning for myCobot}
\label{sec:trajectory-planning-problem}

In this section, we briefly explain how to solve the trajectory planning problem for myCobot as a 3-DOF manipulator.
In the following, assume that the trajectory of the end-effector is given by $C(r)={}^t(C_1(r),C_2(r),C_3(r))\in \Q[r]\times\Q[r]\times\Q[r]$ for $r\in[0,1]$, and the position of the end-effector 
$\p={}^t(p_1,p_2,p_3)$ is given by
\begin{equation}
  \label{eq:trajectory-parameters}
  p_1=C_1(r),\quad p_2=C_2(r),\quad p_3=C_3(r).  
\end{equation}
As a trajectory, we define a line segment given as 
\[
  C_{1}(r) = x_{0}(1 - r) + x_{f}r, \quad
  C_{2}(r) = y_{0}(1 - r) + y_{f}r, \quad
  C_{3}(r) = z_{0}(1 - r) + z_{f}r,
\]
where the initial position of the end-effector is given by ${}^t(x_0,y_0,z_0)$ and the final position of the end-effector is given by ${}^t(x_f,y_f,z_f)$, 
or a natural cubic spline given as
$C_i(r) = a_{i}r^{3} + b_{i}r^{2} + c_{i}r + d_{i}$ for $i=1,2,3$,
where the coefficients $a_i$, $b_i$, $c_i$, and $d_i$ are calculated by the initial position $(x_0, y_0, z_0)$, the final position $(x_3, y_3, z_3)$, and two intermediate positions $(x_1, y_1, z_1)$ and $(x_2, y_2, z_2)$ of the end-effector given in $\R^3$ (see Shirato et al.\ \cite{shi-oka-ter-mik2024}).

Let $\rr$ be a finite monotonically increasing sequence of rational numbers in $[0,1]$ given as
\[
  \rr=\{r_1,\dots,r_m\},\quad 0=r_1<r_2<\cdots<r_m=1.
\]
Then, we solve the trajectory planning problem by solving the IK problem for each point $\p=C(r_i)$ for $i=1,\dots,m$.

\section{Certification of Solutions to the Trajectory Planning Problem for myCobot}
\label{sec:certification-problem}

In this section, we propose a method for certifying solutions to the trajectory planning problem for myCobot using the CGS-QE method \cite{fuk-iwa-sat2015}.
More specifically, as described in the previous section, for a given trajectory $C(r)$ for $r\in[0,1]$, 
we derive the range of $r$ (which is a subset of $[0,1]$)
for which the trajectory planning is feasible, that is, for which the IK problem has a solution.
Note regarding the CGS-QE method that,
whereas their original method has been proposed for quantifier elimination of logical formulas involving constraints in polynomial equations, inequalities, and inequations, we simplify the method to apply it only to constraints in polynomial equations, since the IK equations are given as polynomial equations.

\subsection{The CGS-QE method}
\label{sec:cgs-qe-method}

The theoretical foundation of the CGS-QE method lies in counting the number of real roots of a system of polynomial equations based on observations regarding the multiplication map over the residue class ring of the ideal generated by the system of equations. 
Here, we briefly explain an overview of the method (see Fukasaku et al. \cite{fuk-iwa-sat2015} for details).
Let $R$ be a real closed field, and let $K$ be a computable subfield of $R$.
Let $\bar{X}$ be variables $X_1,\dots,X_n$ and let $I\subset K[\bar{X}]$ be a zero-dimensional ideal. 
Let $V_R(I)$ be the variety of $I$ in $R$.
Then, the residue class ring $K[\bar{X}]/I$ is a finite-dimensional vector space over $K$; let $d$ be the dimension of $K[\bar{X}]/I$ and $\{v_1,\dots,v_d\}$ be a basis of $K[\bar{X}]/I$.
For $h\in K[\bar{X}]/I$ and $i,j$ ($1\leq i,j\leq d$), define a linear map 
$\theta_{h,i,j}:K[\bar{X}]/I\to K[\bar{X}]/I$ by 
$K[\bar{X}]/I\ni f\mapsto hv_iv_jf\in K[\bar{X}]/I$.
Let $q_{h,i,j}$ be the trace of $\theta_{h,i,j}$, and let $M_h^I=(q_{h,i,j})_{i,j}$ be the $d\times d$ matrix (note that $M_h^I$ is symmetric).
Let $\chi_h^I(\lambda)$ be the characteristic polynomial of $M_h^I$,
and $\sigma(M_h^I)$ be the signature of $M_h^I$, which is defined as the difference between the number of positive and negative eigenvalues of $M_h^I$ (note that, since $M_h^I$ is symmetric, all its eigenvalues are real). 
Then, we have the following observations:

\begin{theorem}[\textmd{\cite[Theorem 1]{fuk-iwa-sat2015}}]
   $\sigma(M_{h}^{I}) =
    \#(\{ \bar{c} \in V_{R}(I) \mid h(\bar{c}) > 0 \}) -
    \#(\{ \bar{c} \in V_{R}(I) \mid h(\bar{c}) < 0 \})$.
\end{theorem}

\begin{corollary}[\textmd{\cite[Corollary 2]{fuk-iwa-sat2015}}]
  $\#V_{R}(I) = \sigma(M_{1}^{I})$.
\end{corollary}

In the following, let $R=\R$ and $K=\Q$.

\begin{lemma}[\textmd{\cite[Lemma 12]{fuk-iwa-sat2015}}]
  \label{lem:sign-changes}
  Let $M$ be a symmetric matrix of rational numbers of size $d \times d$, and let $\chi(\lambda)$ be its characteristic polynomial expressed as 
  $\chi(\lambda) = \lambda^d + a_{d-1}\lambda^{d-1} + \cdots + a_0$ and
  $\chi(-\lambda) = (-1)^d\lambda^d + b_{d-1}\lambda^{d-1} + \cdots + b_0$.
  (Note that $b_i=a_i$ for $i$ that is even, and $b_i=-a_i$ for $i$ that is odd.)
  Let $S_+$ be the number of sign changes in the sequence $(1,a_{d-1},\dots,a_0)$, and let $S_-$ be the number of sign changes in the sequence $((-1)^d,b_{d-1},\dots,b_0)$.
  Then, we have 
  \[
  S_+ = \#\{c\in\R\mid c>0\wedge\chi(c)=0\},\quad
  S_- = \#\{c\in\R\mid c<0\wedge\chi(c)=0\}.
  \]
\end{lemma}

\begin{corollary}[\textmd{\cite[Corollary 13]{fuk-iwa-sat2015}}]
  Let $S_+$ and $S_-$ be as in the above lemma. Then, we have 
  $\#V_{R}(I) = \sigma(M_{1}^{I})>0$ if and only if $S_+ \ne S_-$.
\end{corollary}

\begin{definition}[\textmd{\cite[Definition 14]{fuk-iwa-sat2015}}]
  By regarding $a_{d-1},\dots,a_0$ in \Cref{lem:sign-changes} as variables, we can express the formula $S_+ \ne S_-$ as a quantifier-free first-order formula. 
  We denote it by $I_d(a_{d-1},\dots,a_0)$.
\end{definition}

We derive $I_d(a_{d-1},\dots,a_0)$ as follows.
First, let 
$S=\{\bm{\rho}=(\rho_{d-1},\dots,\rho_{0})\mid\rho_{i}\in\{+, 0, -\} \}$
be the set of all possible sign sequences of length $d$.
Then, for each $\bm{\rho}=(\rho_{d-1},\dots,\rho_{0})\in S$, 
let $\bar{\bm{\rho}}=(+,\rho_{d-1},\dots,\rho_{0})$ be the sign sequence by adding $+$ at the beginning of $\bm{\rho}$.
By regarding $\bar{\bm{\rho}}$ as a sequence of signs for the sequence of coefficients 
$(1,a_{d-1},\dots,a_0)$ in \Cref{lem:sign-changes}, 
calculate $S_+$ and $S_-$ for $\bar{\bm{\rho}}$.
Let $S'$ be the set of all $\bm{\rho}\in S$ such that $S_+ \ne S_-$ for $\bar{\bm{\rho}}$.
Then, we have 
\begin{equation}
  \label{eq:I_d}
  I_d(a_{d-1},\dots,a_0) = \bigvee_{\bm{\rho}\in S'}(a_0\;\rho'_1\;0 \wedge\cdots\wedge
   a_i\;\rho'_i\;0 \wedge\cdots\wedge a_{d-1}\;\rho'_{d-1}\;0),
\end{equation}
where
\[
  \rho'_i=
  \begin{cases}
    > & \text{if $\rho_i=+$}, \\
    = & \text{if $\rho_i=0$}, \\
    < & \text{if $\rho_i=-$}.
  \end{cases}
\]

\subsection{Certifying solutions to the trajectory planning problem}
\label{sec:certifying-solutions}

For certifying solutions to the trajectory planning problem, we give the parametric polynomial system 
as follows: for $f_i$ in
\eqref{eq:inverse-kinematics-polynomial-system}, let
\[
  \bar{f}_i(r) =
  \begin{cases}
    \text{$f_i$ with replacing $p_i$ by $C_i(r)$ in \eqref{eq:trajectory-parameters}} & \text{for $1\leq i\leq 3$}, \\
    \text{$f_i$} & \text{for $4\leq i\leq 6$},
  \end{cases}
\]
For the parametric ideal $\bar{I}=\langle \bar{f}_1(r),\dots,\bar{f}_6(r)\rangle \subset \R(r)[c_1,s_1,c_2,s_2,c_3,s_3]$ with parameter $r$, 
calculate the first-order fomula $I_d(a_{d-1},\dots,a_0)$ in \eqref{eq:I_d} for the characteristic polynomial of $M_1^{\bar{I}}$ as described in \Cref{sec:cgs-qe-method}.
Note that,
in $I_d(a_{d-1},\dots,a_0)$,
$a_{d-1},\dots,a_0$ are 
rational functions in $r$, denoted by $a_{d-1}(r),\dots,a_0(r)$.
Thus, $I_d$, denoted as $I_d(a_{d-1}(r),\dots,a_0(r))$, is a first-order formula with variable $r$.

Here, let $\varphi(r)=I_d(a_{d-1}(r),\dots,a_0(r))\wedge \cF(\cS)$, where
$\cF(\cS)$ is defined as follows: 
$\cS=V_{\C}(I_1)\setminus V_{\C}(I_2)$ is the segment of 
the CGS for $\bar{I}$ that is chosen for a certain value of $r$, where $I_1$ and $I_2$ are ideals in $\Q[r]$, given as $I_1=\langle g_1(r),\dots,g_k(r)\rangle$ and $I_2=\langle h_1(r),\dots,h_l(r)\rangle$ for some polynomials $g_i(r)$ and $h_j(r)$ in $\Q[r]$,
for $i=1,\dots,k$ and $j=1,\dots,l$.
Then, the defining formula $\cF(\cS)$ of $\cS$ is defined as 
$g_1(r)=0\wedge\cdots\wedge g_k(r)=0\wedge (h_1(r)\ne 0\vee\cdots\vee h_l(r)\ne 0)$
\cite[Def.\ 15]{fuk-iwa-sat2015}.
The algorithms in the CGS-QE method \cite[Alg.\ 1 and 2]{fuk-iwa-sat2015} tells us that
the formula $\varphi(r)$ is equivalent to the formula $\exists c_1\exists s_1\exists c_2\exists s_2\exists c_3\exists s_3(\bar{f}_1(r)=0\wedge\cdots\wedge \bar{f}_6(r)=0)$
for all $r\in\cS$.
Furthermore, the formula with rational functions in $r$ which appears in $I_d(a_{d-1}(r),\dots,a_0(r))$ can be transformed into formulas with only polynomials in $r$, as follows: 
for $a_i(r)=n_i(r)/d_i(r)$ ($i=0,\dots,d-1$), where $n_i(r)$ and $d_i(r)$ are coprime polynomials in $\Q[r]$, $d_i(r)$ is a product of the leading coefficients of elements in the Gröbner basis for $\bar{I}$ which do not vanish on $\cS$. Thus, the equation and the inequalities in $a_i(r)$ can be transformed into the equations and the inequalities in $n_i(r)$ and $d_i(r)$ as
\begin{align*}
  a_i(r)=0 &\Leftrightarrow n_i(r)=0, \\
  a_i(r)>0 &\Leftrightarrow (n_i(r)>0\wedge d_i(r)>0) \vee (n_i(r)<0\wedge d_i(r)<0),\\
  a_i(r)<0 &\Leftrightarrow (n_i(r)>0\wedge d_i(r)<0) \vee (n_i(r)<0\wedge d_i(r)>0),
\end{align*}
(note that $a_i(r)\ne 0$ can be transformed into $a_i(r)>0\vee a_i(r)<0$).

As a result, $\varphi(r)$ can be transformed into $\psi(r)$, a disjunctive normal form of first-order formulas consisting of \emph{polynomial} equations and inequalities in $r$.
Now, let $[\gamma,\delta]$, where $\gamma,\delta\in\Q$, be a real closed interval with 
$-\infty<\gamma<\delta<\infty$.
Then, we determine the subset $\cR\subset [\gamma,\delta]$ in which $\psi(r)$ holds for any 
$r\in\cR$, as follows.

Let $F$ be the set of all polynomials in $r$ that appear in $\psi(r)$, and let 
$\bar{F}$ be the set of coprime factors of the polynomials in $F$.
Let $\{\alpha_1,\dots,\alpha_l\}$ be the set of distinct real roots of all the polynomials $q\in\bar{F}$ in 
$[\gamma,\delta]$ satisfying $\gamma<\alpha_1<\cdots<\alpha_l<\delta$ (in the case there exists $q\in\bar{F}$ satisfying $q(\gamma)=0$ 
(or $q(\delta)=0$, respectively), slightly shift $\gamma$ to $\gamma-\varepsilon$ (or $\delta$ to 
$\delta+\varepsilon$, repetitively), with $\varepsilon$ a small rational number). 
For $i=1,\dots,l$, calculate isolating intervals \cite{bas-pol-roy2006} of $\alpha_i$ in $[\gamma,\delta]$, which are given as $(\gamma_i,\delta_i]$, where $\gamma_i$ and $\delta_i$ are rational numbers such that 
\begin{multline}
  \label{eq:isolating-intervals}
  \gamma<\gamma_1<\alpha_1<\delta_1<\gamma_2<\alpha_2<\delta_2<\cdots<\delta_{i-1}<\gamma_i<\alpha_i<\delta_i
  \\ < \gamma_{i+1}<\alpha_{i+1}<\delta_{i+1}<\cdots
  <\delta_{l-1}<\gamma_l<\alpha_l<\delta_l<\delta.  
\end{multline}
By setting $\bar{\alpha}_i=((\gamma_i,\delta_i],q_i)$, we denote the pair of the isolating interval $(\gamma_i,\delta_i]$ and the corresponding polynomial $q_i\in\bar{F}$ such that $q_i(\alpha_i)=0$, 
and regard it as the root $\alpha_i$.
Then, for each root $\alpha_i$ and its isolating interval $\bar{\alpha}_i=((\gamma_i,\delta_i],q_i)$, we have the following observations.
\begin{enumerate}
  \item The boolean value of $\psi(r)$ may change only at $r=\alpha_i$
  since the sign of a polynomial changes only at its roots.
  Thus, $\psi(r)$ is true for any $r\in(\alpha_i,\alpha_{i+1})$ if and only if $\psi(\delta_i)$ is true.
  \item The boolean value of $\psi(\alpha_i)$ can be determined as follows: 
  for a polynomial $p(r)$ appearing in $\psi(r)$,
  $p(\alpha_i)=0$ if and only if $q_i(r)$ divides $p(r)$.
  If $q_i(r)$ does not divide $p(r)$, then
  the sign of $p(\alpha_i)$ is the same as the sign of $p(\delta_i)$.
\end{enumerate}

By applying the above observations to all roots $\alpha_i$ and their isolating intervals $\bar{\alpha}_i$, we can determine the subset $\cR\subset [\gamma,\delta]$ in which $\psi(r)$ holds for any $r\in\cR$, summarized in \Cref{alg:detecting-range}.



\bibliographystyle{splncs04}
\bibliography{icms-2026-nakai-terui-mikawa}


\begin{algorithm}[tb]
  \caption{Deriving of the range of variables satisfying a disjunctive normal form}
  \label{alg:detecting-range}
  \begin{algorithmic}[1]
      \Input $\psi(r)$: a disjuncive normal form, $[\gamma,\delta]\subset\R$ with 
      $\gamma,\delta\in\Q$ such that $\gamma<\delta$, $\epsilon>0$: a small rational number (say, $\epsilon=1/100$)
      \Output $\cR\subset[\gamma,\delta]$: the subset such that $\psi(r)$ holds for any $r\in\cR$
      \State $F\gets\text{(the set of all polynomials in $r$ that appear in $\psi(r)$)}$
      \State $\bar{F}\gets\text{(the set of coprime factors of the polynomials in $F$)}$
      \ForAll{$f \in \bar{F}$}
        \If{$f(\gamma)=0$}
          \State $\gamma\gets\gamma-\epsilon$
        \EndIf
        \If{$f(\delta) = 0$}
          \State $\delta\gets\delta+\epsilon$
        \EndIf
      \EndFor
      \State $\{\bar{\alpha}_{1}=((\gamma_1, \delta_1],f_1),\dots,
      \bar{\alpha}_{l}= ((\gamma_l, \delta_l], f_{l})\}\gets$
      (the set of isolating intervals satisfying \eqref{eq:isolating-intervals} for the set of distinct real roots $\{\alpha_1,\dots,\alpha_l\}\subset[\gamma,\delta]$ of all the polynomials $f\in\bar{F}$)
      \Comment{Regard $\bar{\alpha}_{i}$ as the root $\alpha_i$ of $f_{i}$ in $(\gamma_i, \delta_i]$}
      \State $\mathcal{R}\gets\emptyset$
      \If{$\psi(\gamma)$}
          \State $\mathcal{R} \gets \mathcal{R} \cup [\gamma, \alpha_{1})$
      \EndIf
      \For{$i = 1, \dots, l-1$}
          \If{$\psi(\alpha_{i})$}
              \State $\mathcal{R} \gets \mathcal{R} \cup \{\alpha_{i}\}$
          \EndIf
          \If{$\psi(\delta_i)$}
              \State $\mathcal{R} \gets \mathcal{R} \cup (\alpha_{i}, \alpha_{i+1})$
          \EndIf
      \EndFor
      \If{$\psi(\delta)$}
          \State $\mathcal{R} \gets \mathcal{R} \cup (\alpha_{l}, \delta]$
      \EndIf
      \State \textbf{return} $\mathcal{R}$
  \end{algorithmic}
\end{algorithm}

\end{document}